\documentclass{article}

 \usepackage[preprint]{neurips_2026}

\usepackage[utf8]{inputenc} 
\usepackage[T1]{fontenc}    
\usepackage{hyperref}       
\usepackage{url}            
\usepackage{booktabs}       
\usepackage{amsfonts}       
\usepackage{nicefrac}       
\usepackage{microtype}      
\usepackage{xcolor}         

\usepackage{graphicx}
\usepackage{multirow} 
\usepackage{enumitem}
\usepackage{tabulary}
\usepackage{colortbl}
\usepackage{wrapfig}
\usepackage{amsmath, amssymb}
\usepackage{algorithm}
\usepackage{algpseudocodex}
\usepackage{diagbox}
\usepackage{pifont}
\newcommand{\cmark}{\ding{51}}
\newcommand{\xmark}{\ding{55}}
\usepackage{graphicx}
\usepackage{subcaption}
\usepackage{titlesec}
\titlespacing*{\section}{0pt}{0.35em}{0.2em}
\titlespacing*{\subsection}{0pt}{0.3em}{0.1em}
\titlespacing*{\subsubsection}{0pt}{0.25em}{0.1em}

\usepackage{booktabs}
\usepackage{tabularx}
\usepackage{makecell}

\newcommand{\cutabstractup}{\vspace*{-0.13in}}
\newcommand{\cutabstractdown}{\vspace*{-0.1in}}

\newcommand{\cutsectiondown}{\vspace*{-0.07in}}

\newcommand{\cutsubsectiondown}{\vspace*{-0.04in}}

\title{SECOND-Grasp: Semantic Contact-guided \\ Dexterous Grasping}

\author{
  \textbf{Han Yi Shin}\textsuperscript{1}\thanks{Equal contribution.} \quad
  \textbf{Heeju Ko}\textsuperscript{1}\footnotemark[1] \quad
  \textbf{Jaewon Mun}\textsuperscript{1} \quad
  \textbf{Qixing Huang}\textsuperscript{2} \quad
  \textbf{Jaehyeok Lee}\textsuperscript{1} \\
  \vspace{0.2cm}
  \textbf{Sung June Kim}\textsuperscript{1} \quad
  \textbf{Honglak Lee}\textsuperscript{3} \quad
  \textbf{Sujin Jang}\textsuperscript{4} \quad
  \textbf{Sangpil Kim}\textsuperscript{1}\thanks{Corresponding author.} \\
  \\
  \makebox[\textwidth][c]{
    \textsuperscript{1}Korea University \:
    \textsuperscript{2}University of Texas at Austin \:
    \textsuperscript{3}University of Michigan \:
    \textsuperscript{4}Hanyang University ERICA
  }
}

\begin{document}

\maketitle

\begin{abstract}
\cutabstractup

Achieving reliable robotic manipulation, such as dexterous grasping, requires a synergy between physically stable interactions and semantic task guidance, yet these objectives are often treated as separate, disjoint goals. 
In this paper, we investigate how to integrate dexterous grasping techniques, i.e., physically stable grasps for object lifting and language-guided grasp generation, to achieve both physical stability and semantic understanding. 
To this end, we propose \textbf{SECOND-Grasp} (SEmantic CONtact-guided Dexterous Grasping), a unified framework that enables robotic hands to dynamically adjust grasping strategies based on semantic reasoning while ensuring physical feasibility.
We begin by obtaining coarse contact proposals through vision-language reasoning to infer where contacts should occur based on object properties, followed by segmentation to localize these regions across views.
To further ensure consistency across multiple viewpoints, we introduce Semantic-Geometric Consistency Refinement (SGCR), which refines initial contact predictions by enforcing semantic consistency across views and removing geometrically invalid regions, yielding reliable 3D contact maps.
Then, we derive a feasible hand pose for each contact map via inverse kinematics, generating a supervision signal for policy learning.
Our approach, trained on DexGraspNet, consistently outperforms baselines in lifting success rate on both seen and unseen categories, achieving 98.2\% and 97.7\%, respectively, while also improving intent-aware grasping by 12.8\% and 26.2\%. 
We further show promising results on additional datasets and robotic hands, including Shadow Hand and Allegro Hand.
\cutabstractdown
\end{abstract}
\vspace{-0.3em}
\section{Introduction}
\cutsectiondown

Dexterous grasping has been widely studied as a fundamental capability for robotic manipulation, with the primary goal of achieving physically stable object interactions~\cite{wang2022dexgraspnet,xu2023unidexgrasp, wan2023unidexgrasp++,xu2024dexterous,zhong2025dexgrasp}. Recent advances have focused on generating multi-fingered hand configurations~\cite{xu2023unidexgrasp, xu2024dexterous, ma2025contact} that satisfy force closure, collision avoidance, and robust object lifting. However, these approaches largely operate in a pose-centric manner, with limited consideration of high-level task intent or semantic guidance. 
In contrast, vision-language reasoning-based manipulation for multi-fingered hands~\cite{he2025dexvlg, zhong2026dexgraspvla, lee2026dexter} has emerged as a promising direction for leveraging semantic cues in action generation, but it often focuses on producing plausible hand poses without explicitly ensuring physical feasibility~\cite{lee2026dexter, li2024multi}.

\begin{figure}[t]
  \centering
  \includegraphics[width=\textwidth]{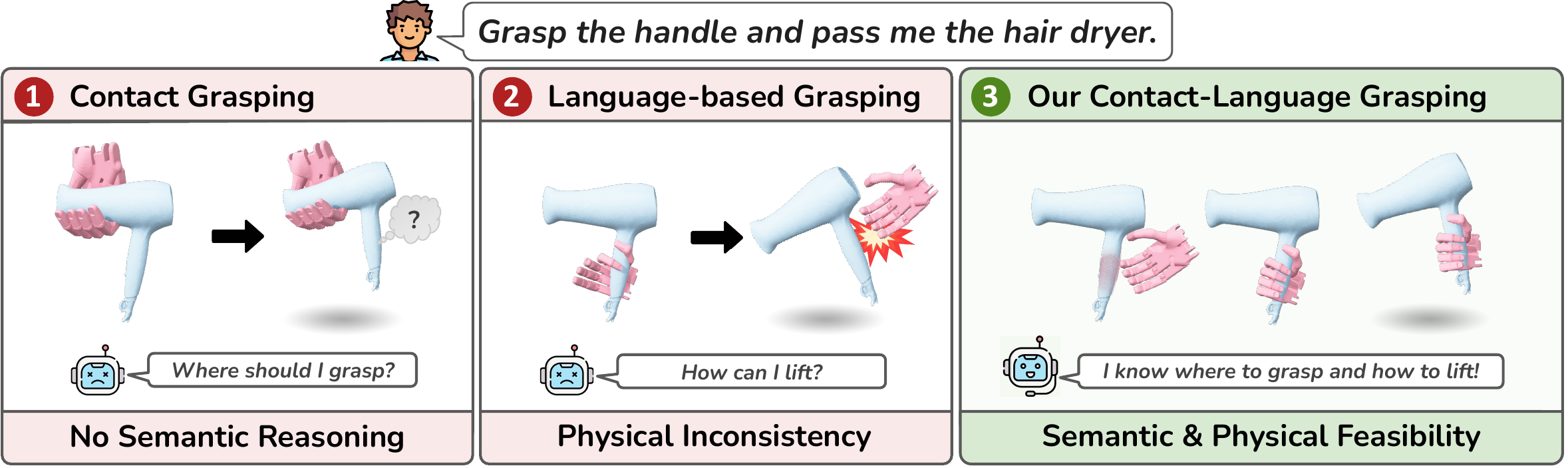}
    \caption{Contact-based grasping ensures physical feasibility but lacks semantic reasoning, while language-based grasping captures intent but often fails to lift. 
    SECOND-Grasp bridges both, enabling grasps that are semantically meaningful and physically stable.}
  \label{fig:teaser}
\vspace{-1.0em}
\end{figure}


This disconnection raises two fundamental challenges: without semantic grounding, grasp strategies cannot adapt to task-specific requirements~\cite{li2024shapegrasp, wei2024grasp, lee2026dexter}, and without explicit contact modeling, they fail to ensure physically stable execution in practice~\cite{liu2023contactgen,liu2021synthesizing, wang2022dexgraspnet}. 
Existing methods address these as separate problems, yet bridging the two remains largely unexplored. 
We argue that truly capable dexterous manipulation requires bridging this gap through explicit contact-level reasoning that jointly grounds semantic intent and physical feasibility.

To address this, we propose \textbf{SECOND-Grasp} (SEmantic CONtact-guided Dexterous Grasping), a unified framework that explicitly models contact-level interactions to couple semantic understanding with physical feasibility in grasp planning.
These contact regions correspond to semantic object parts (e.g., handle, body, base), spatially grounding semantic intent to guide grasp planning.
Rather than directly predicting grasps from observations, \textbf{SECOND-Grasp} first infers where meaningful contacts should occur, and then uses these contacts to guide robot hand grasping.
To this end, it leverages the semantic priors of vision-language models to reason about object-relevant contact regions, and grounds these predictions through segmentation across multi-view observations.

However, while vision-language and segmentation models provide strong semantic priors, their predictions are made independently per view and thus may still exhibit misalignment between semantic agreement and geometric consistency across views. We therefore introduce Semantic-Geometric Consistency Refinement (SGCR), which enforces cross-view consistency through two sequential stages: (1) a Semantic Refinement Stage that assigns higher confidence to regions consistently predicted across multiple views; and 
(2) a Geometric Refinement Stage that enforces geometric consistency of the contact map using a local convexity-based criterion, which resolves surface ambiguity by verifying whether predictions lie on the same surface region.
Together, these two stages yield contact maps that are both semantically grounded and geometrically coherent, enabling them to capture diverse semantic and geometric interaction intents across object properties.

To use these contact intentions for policy learning, we need to translate them into hand-level guidance that supports physical execution.
A common strategy is to recover a grasp pose from the contact map using learned grasp generators~\cite{weng2024dexdiffuser, lu2024ugg, wu2025fastgrasp}.
However, these models are trained on fixed hand-object datasets and thus inherit hand- and object-specific priors, which can limit generalization to new hands or objects. To address this issue, we adopt contact-based optimization methods~\cite{liu2023contactgen, zhao2024graingrasp, zhang2026cadgrasp}, which provide an effective way to convert contact representations into hand configurations. However, unlike prior contact-based pipelines that use the recovered configuration as the final grasp, we use analytic inverse kinematics to derive pseudo pose priors from the refined contact maps for policy learning.
Together with the refined contact maps, these priors guide policy learning toward semantically aligned and stable grasping behaviors across diverse objects and robotic hands.


Experimental results demonstrate that our approach, trained on DexGraspNet dataset~\cite{zhang2024dexgraspnet}, outperforms existing grasping methods~\cite{xu2023unidexgrasp, wan2023unidexgrasp++, wang2025unigrasptransformer, yuan2025demograsp, mao2025universal} in object lifting success rate, achieving 98.2\% and 97.7\% on seen and unseen categories, respectively. 
Beyond general grasping performance, we evaluate intention alignment success rate, observing improvements of 12.8\% and 26.2\% for seen and unseen categories.
We further validate the transferability of our framework to different hand embodiments using the Shadow Hand~\cite{shadowhand} and Allegro Hand~\cite{allegrohand}, and evaluate its generalizability across diverse object datasets.

Our contributions are summarized as follows:
\begin{itemize}[leftmargin=*]
    \item SECOND-Grasp, a grasping framework that bridges semantic intent with execution by predicting semantically meaningful contact regions and using them to guide physically stable grasping.
    \item Semantic-Geometric Consistency Refinement (SGCR), a two-stage module that enforces cross-view consistency via semantic scoring and local convexity validation, producing consistent contact maps.
    \item A learning pipeline that leverages pseudo pose priors derived from contact maps, enabling robust grasping performance across diverse objects, tasks, and robotic hands.
\end{itemize}

\begin{figure}[t]
\begin{center}
\includegraphics[width=\linewidth]{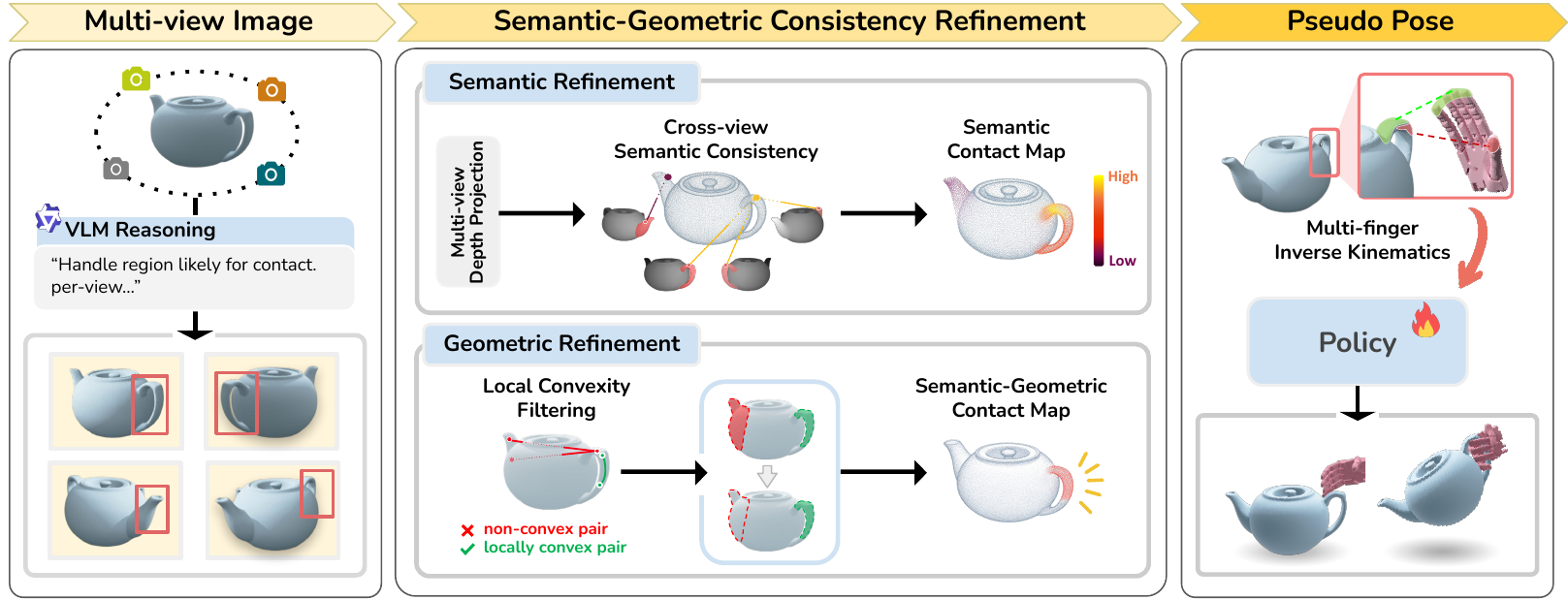}
    \end{center}
    \caption{\textbf{Overview.} 
    Given multi-view observations, \textbf{SECOND-Grasp} first infers semantic contact regions using vision-language reasoning (Section~\ref{sec:contact-region-proposal}). These 2D proposals are then projected into 3D and refined through Semantic-Geometric Consistency Refinement, which enforces cross-view semantic consistency and geometric consistency based on local convexity (Section~\ref{sec:semantic-geometric-consistency-refinement}).
    The refined semantic-geometric contact map provides inverse-kinematics-based pseudo pose guidance (Section~\ref{sec:ik_step}) for policy learning (Section~\ref{policy_learning}). See individual sections for technical details.
    }
    \label{fig:main}
\vspace{-1.5em}
\end{figure}


\section{Related Work}
\cutsectiondown

\subsection{Dexterous Grasping}
\cutsubsectiondown

Dexterous grasping aims to achieve stable object grasping and manipulation with multi-fingered robotic hands.
To enable scalable learning of dexterous grasping, prior methods have constructed large-scale datasets that pair objects with physically dexterous hand poses~\cite{wang2022dexgraspnet,li2023gendexgrasp, liu2024realdex, zhong2025dexgrasp, taheri2020grab}.
Building on large-scale grasp datasets, policy-learning methods train dexterous hands for object lifting across diverse objects and hand embodiments~\cite{xu2023unidexgrasp,wan2023unidexgrasp++,resdex,yuan2025demograsp,mao2025universal}.
However, these methods primarily focus on physical feasibility or lifting success, without explicitly considering whether the resulting contacts align with semantically meaningful object regions.
Recent language-guided dexterous grasping methods incorporate task intent or object usage into grasp generation~\cite{wei2024grasp,wei2025afforddexgrasp,lee2026dexter, jian2025g}, 
but remain focused on grasp pose synthesis and do not address physically executable lifting. 
Our work bridges these two directions by grounding dexterous policy learning in contact regions inferred through 2D VLM reasoning, enabling grasps that are semantically meaningful and physically executable.

\subsection{2D-to-3D Affordance Grounding}
\cutsubsectiondown

Affordance grounding aims to identify object regions that enable specific interactions~\cite{gibson1979theory, do2018affordancenet, qian2024affordancellm, li2024one, zhu2025grounding, moon2025selective}.
Understanding affordances in 3D is essential for embodied interaction and robotic manipulation, as it enables agents to reason about how objects can be used in the physical world~\cite{deng20213d, qian2024affordancellm}. However, object affordance is not always determined by geometry alone, as functional regions often depend on how an object is used. This has motivated approaches to incorporate 2D interaction or semantic cues into 3D affordance grounding. IAGNet~\cite{yang2023grounding} and MIFAG~\cite{gao2025learning} use 2D human-object interaction images to guide 3D affordance prediction, while CAST~\cite{huang2025unlocking} transfers semantic features from 2D vision foundation models into 3D representations. Together, these methods demonstrate that 2D semantic knowledge provides effective priors for localizing functional regions in 3D. Our work builds on this insight, applying 2D VLM reasoning to infer object-specific contact regions and refining them into 3D contact maps for dexterous grasping.

\subsection{Grasp Pose Generation}
\cutsubsectiondown


Dexterous grasp generation methods commonly synthesize hand configurations from object observations using learned generative models or analytical solvers.
Generative methods learn grasp distributions conditioned on the features of the object via diffusion-based modeling~\cite{weng2024dexdiffuser, lu2024ugg, wu2025fastgrasp}.
Although effective in generating diverse and plausible grasps, these models often learn hand-specific priors tied to training embodiments and object distributions, which can limit transferability to novel hands or objects.
Analytical methods instead recover hand configurations from geometric or kinematic constraints without learning a grasp distribution~\cite{turpin2022grasp, turpin2023fast, liu2021synthesizing, zurbrugg2025graspqp, murray2017mathematical, siciliano2009robotics, buss2004introduction}.
Within analytical approaches, contact-based optimization methods use predicted contacts to recover hand configurations, demonstrating contact as an effective intermediate representation for grasp synthesis~\cite{liu2023contactgen, zhao2024graingrasp, zhang2026cadgrasp}.
Our work builds on these perspectives, but rather than treating the recovered configuration as the final grasp, we use it as a pseudo-pose prior to guide policy learning toward executable grasping behavior.

\section{Preliminaries}
\cutsectiondown

Dexterous grasping requires training a policy for a multi-fingered robot to interact with diverse objects. We formulate the task as a Partially Observable Markov Decision Process (POMDP) with timesteps  $t = 0, \dots, T$. For each timestep, the environment evolves according to state:

\begin{equation}
s_t = (x_t^{hand}, x_t^{obj}, x^{geo}),
\end{equation}
where $x_t^{hand}$ denotes robot proprioceptive states (e.g., joint configurations and fingertip states), $x_t^{obj}$ denotes object pose and velocities, and $x^{geo}$ encodes object geometric information. Due to partial observability, the agent instead receives an observation:
\begin{equation}
o_t = (x_t^{hand}, v_t^{img}, v_t^{point}),
\end{equation}
which is generated from the underlying state via an observation function $P(o_t | s_t)$. Here, $v_t^{img}$ denotes object-centric images from a third-person view and $v_t^{point}$ denotes partial point cloud observations.

Using full access to the underlying state in simulation, we first train a privileged state-based policy $\pi_{\mathrm{state}}(a_t \mid s_t)$. We then distill the policy into a vision-based policy $\pi_{\mathrm{vision}}(a_t \mid o_t)$ that could operate under partial observability in real-world settings. For both policies, the action is parameterized by:  
\begin{equation}
a_t = (\Delta w_t, \Delta \phi_t, \Delta \theta_t),
\end{equation}
where $\Delta w_t \in \mathbb{R}^3$ and $\Delta \phi_t \in \mathbb{R}^3$ denote the translational and rotational increases in the wrist, respectively, and $\Delta \theta_t \in \mathbb{R}^{d_\theta}$ denotes the incremental joint configurations for the actuated hand joints. Here, $d_\theta$ depends on the robotic hand embodiment (see Appendix~\ref{app:transfer_allegro_hand}). Both policies are trained to maximize the expected discounted return $\mathbb{E}[\sum_{t=0}^{T-1} \gamma^t {r}_t]$, where $\gamma \in (0,1)$ is the discount factor.

\section{Method}
\cutsectiondown

Our SECOND-Grasp enables robotic hands to grasp objects in a semantically meaningful and physically plausible manner. To achieve this, we first generate a 3D contact map that represents object-specific graspable regions, and then infer a corresponding initial hand pose via inverse kinematics. 
We subsequently use the contact map and inverse-kinematics-derived pose prior to guide policy learning, enabling semantically aligned and physically stable grasping behaviors.
The overview of SECOND-Grasp is illustrated in Figure~\ref{fig:main}.

\subsection{Semantic Contact Region Proposal}\label{sec:contact-region-proposal}
\cutsubsectiondown

Identifying semantic contact regions solely from 3D geometry is ambiguous, as graspable parts are defined not just by shape but by functional intent—how an object is meant to be used. This motivates us to leverage 2D vision-language reasoning as a semantic prior to guide contact region proposal.

Given multi-view RGB images of an object $\{\mathcal{I}_i\}_{i=1}^{I}$ from $I=4$ fixed viewpoints (front, back, left, right), we use a pretrained large vision-language model~(VLM)~\cite{bai2025qwen3} to extract an initial set of $K$ graspable regions $\{g_k\}_{k=1}^{K}$ (e.g., handle, body, base) and their 2D bounding boxes $\{b_k\}_{k=1}^{K}$. Additionally, each bounding box is associated with a confidence score $c_i^{(k)} \in [0,1]$, which indicates the VLM's estimated likelihood that the bounding box contains the $k$-th graspable region. To achieve a precise pixel-level localization, we apply a segmentation model~\cite{ravi2024sam} $\mathcal{F}_\mathrm{seg}$ within each bounding box and obtain a binary mask $M_i^{(k)} = \mathcal{F}_\mathrm{seg}(I_i, b_i^{(k)})$. The mask is further filtered by valid object regions from the rendered depth map to yield $\tilde{M}_i^{(k)}$.




\subsection{Semantic-Geometric Consistency Refinement} 
\label{sec:semantic-geometric-consistency-refinement}
\cutsubsectiondown

While the 2D VLM provides high-level semantic contact region proposals, accurate robotic grasping requires finer refinement in the 3D space. Therefore, we introduce a Semantic-Geometric Consistency Refinement (SGCR) module to refine these coarse proposals into precise 3D contact regions that are consistent with both semantic and geometric cues.


\subsubsection{Contact Confidence Initialization}
The confidence score $c_n^{(k)}$ assigned by the VLM reflects the model’s semantic confidence in the presence of the k-th graspable region, but do not account for precise localization quality. To address this, we re-weight the raw confidence $c_i^{(k)}$ by the valid object region ratio $\rho_i^{(k)}$, which is obtained by the area of $\tilde{M}_i^{(k)}$ over the predicted bounding box:
\begin{equation}
\bar{c}_i^{(k)} = \sigma(\rho_i^{(k)}) \, c_i^{(k)},
\end{equation}
where $\sigma(\cdot)$ is a sigmoid function for down-weighting predictions with low overlap. The initial per-view confidence map is then constructed as $S_i^{(k,0)}(p) = \bar{c}_i^{(k)} \cdot \tilde{M}_i^{(k)}(p)$, where $p$ denotes a pixel location, assigning the calibrated score to each valid mask pixel. 





\subsubsection{Cross-View Semantic Refinement}
Regions that are consistently predicted across multiple views better correspond to semantically expressive parts than those predicted from a single view.
We therefore assign higher confidence for such cross-view consistent predictions.

Let $\mathcal{N}(i)$ denote the set of neighboring views of source view $i$. For each pixel $p$ in $\tilde{M}_i^{(k)}$, we back-project $p$ into 3D space and re-project it onto each neighboring view $j \in \mathcal{N}(i)$.
We verify whether the re-projected point is geometrically consistent and satisfies a depth consistency, which is measured by the discrepancy between the actual and the re-projected depth:
\begin{equation}
\label{eq:depth_consistency}
\delta_{i\to j}(p) = \left| D_{i\to j}(p) - z_{i\to j}(p) \right|,
\end{equation}
where $D_{i\to j}(p)$ is the actual depth of the reference pixel point $p$ in view $j$, and $z_{i\to j}(p)$ is the re-projected depth of pixel point $p$ in view $j$. We consider cross-view predictions to be consistent if $\delta_{i\to j}(p)$ is less than a predefined threshold $\tau$ and the re-projected point lies within $\tilde{M}_j^{(k)}$. We then accumulate this cross-view support to refine the source-view confidence:

\begin{equation}
S_i^{(k,\mathrm{ref})}(p)
=
S_i^{(k,0)}(p)
+
\sum_{j \in \mathcal{N}(i)}
\alpha \, \bar{c}_j^{(k)} \,
\mathbf{1}
\big(
\delta_{i \to j}(p) < \tau
\;\land\;
p_{i \to j} \in \tilde{M}_j^{(k)}
\big),
\end{equation}
where $\alpha$ is a scaling factor to moderate the contribution of cross-view agreement, preventing it from overwhelming initial confidence.
The refined maps are then normalized globally across all views by their maximum value to yield $S_i^{(k,\mathrm{norm})}$, which serves as semantic guidance for the subsequent 3D geometric refinement step.
We denote the normalized confidence maps as $\mathcal{S}^{(k,\mathrm{norm})} = \{ S_i^{(k,\mathrm{norm})} \}_{i=1}^I$.


\subsubsection{3D Geometric Refinement via Local Convexity}
\cutsubsectiondown

Refined maps $S^{(k,\mathrm{norm})}$ capture semantic consistency in all views, but do not enforce geometric coherence at the surface level of the object.
As a result, predicted regions may be overly dispersed or fragmented into multiple disconnected components, resulting in ambiguous contact targets for the robotic hand. 
To address this, we propose a novel local convex-pair refinement module that explicitly enforces surface-level geometric coherence, consolidating fragmented predictions into spatially consistent and physically meaningful contact regions.

We first lift the multi-view maps $S^{(k,\mathrm{norm})}$ into 3D by back-projecting each view into object space using the corresponding depth maps, resulting in a set of 3D points $\mathcal{Q}^{(k)} \subset \mathcal{O}$, where $\mathcal{O}$ denotes the object surface. Here, each point is associated with a confidence score inherited from $S^{(k,\mathrm{norm})}$. Then, we define a high-confidence seed region $\mathcal{Q}_{\mathrm{seed}}^{(k)}$ as the top 10\% of points in $\mathcal{Q}^{(k)}$ by confidence scores.

Starting from the seed region $\mathcal{Q}_{\mathrm{seed}}^{(k)}$, we identify points on the same object surface via local-convexity consistency. 
A pair of surface points $(q_1, q_2) \in \mathcal{O} \times \mathcal{O}$ is defined as a locally convex pair if the line segment connecting them lies entirely within the volume of the object:
\begin{equation}
\mathcal{C}(\mathcal{O})
=
\left\{
(q_1, q_2)\in \mathcal{O}\times\mathcal{O}
\;\middle|\; 
(1-\lambda) q_1 + \lambda q_2 \in \mathrm{Vol}(\mathcal{O}),\ \forall \lambda \in [0,1]
\right\}.
\label{eq:local_convex_pair}
\end{equation}

where $\mathrm{Vol}(\mathcal{O})$ denotes the volume enclosed by the object surface $\mathcal{O}$.
Then, we retain a point $q \in \mathcal{Q}^{(k)}$ in the contact map $\mathcal{Q}_{\mathrm{final}}^{(k)}$ if it satisfies local-convexity consistency with any seed point $q_s \in \mathcal{Q}_{\mathrm{seed}}^{(k)}$:

\begin{equation}
\mathcal{Q}_{\mathrm{final}}^{(k)}
=
\mathcal{Q}_{\mathrm{seed}}^{(k)}
\cup
\left\{
q \in \mathcal{Q}^{(k)}
\;\middle|\;
\exists\; q_s \in \mathcal{Q}_{\mathrm{seed}}^{(k)},
(q, q_s)\in \mathcal{C}(\mathcal{O})
\right\}.
\end{equation}

Because candidate points are validated directly against the seeds, rather than through iterative propagation or global expansion, this formulation defines local convexity in a seed-centered, pairwise manner. A candidate point is retained only if it forms a locally convex pair with at least one seed, which naturally restricts the accepted region around each seed to a compact local neighborhood, such as a cylinder-like subset of a possibly curved cup handle. By taking the union of these seed-centered neighborhoods, the semantically grounded seed region is expanded into a geometrically coherent contact map without requiring the entire semantic part to be globally convex.

\subsection{Hand Pose Prior via Inverse Kinematics}
\cutsubsectiondown
\label{sec:ik_step}

To bridge the refined contact map with policy learning, we derive a physically feasible pseudo hand pose through inverse kinematics~(IK). 
Given the refined contact map $\mathcal{Q}_{\mathrm{final}}^{(k)}$, we estimate its principal axis using PCA~\cite{wold1987principal} and partition the contact points into a thumb region $\mathcal{Q}_{\mathrm{thumb}}^{(k)}$ and a non-thumb region $\mathcal{Q}_{\mathrm{non\text{-}thumb}}^{(k)}$, encouraging an oppositional grasp structure. 
Each robot finger $f\in F$ is then assigned a target contact region $\mathcal{R}_f^{(k)}$, where $F$ denotes the set of fingers of the target hand and may vary across embodiments.

Let $h=(w,\phi,\theta)$ denote the hand configuration, where $w\in\mathbb{R}^3$ and $\phi\in\mathbb{R}^3$ are the wrist position and orientation, and $\theta\in\mathbb{R}^{d_\theta}$ is the joint configuration. 
For each finger $f$, let $\xi_f(h)$ be the fingertip position induced by $h$. 
We obtain the pseudo hand pose by minimizing the fingertip-to-region distance:
\vspace{-0.1em}
\begin{equation}
\min_{h}
\sum_{f \in F}
\min_{y \in \mathcal{R}_f^{(k)}}
\left\|
\xi_f(h) - y
\right\|_2^2.
\label{eq:ik_objective}
\end{equation}
\vspace{-0.5em}

This objective is solved with a Jacobian-based damped least-squares IK solver~\cite{buss2004introduction}. 
In practice, we use 12 iterations of IK to obtain an efficient contact-aligned pseudo pose; implementation details are provided in Appendix~\ref{app:ik_solver}.

\subsection{Policy Learning}
\label{policy_learning}
\cutsubsectiondown


The joint configuration $\theta^*$ extracted from the inverse kinematics(IK) solution serves as a pose prior for guiding policy learning.
Together with the refined contact map, this prior shapes the reward to encourage semantically aligned and physically stable grasping behaviors.
The contact map and pseudo pose are used as reward guidance, allowing semantic contact intent to be distilled into the learned policy.
We train $\pi_{\text{state}}$ using reinforcement learning with:

\vspace{-0.5em}
\begin{equation}
\label{eq:policy_reward}
r = r_{\mathrm{contact}} + r_{\mathrm{pose}} + r_{\mathrm{task}},
\end{equation}
\vspace{-0.5em}

where $r_{\text{contact}}$ encourages fingertips to approach the refined contact region, $r_{\text{pose}}$ penalizes deviation from the IK-derived pseudo pose $\theta^*$, and $r_{\text{task}}$ follows standard lifting and goal-distance objectives~\cite{xu2023unidexgrasp}.
At test time, the trained policy predicts actions directly from its observations without requiring contact maps, pseudo poses, or inverse kinematics.

\section{Experiments}
\cutsectiondown

Our experiments aim to evaluate four aspects of the proposed method: 
(1) grasp success and generalization to unseen object instances and categories,
(2) intent-aware grasping, including contact alignment and grasp-style diversity,
(3) the effectiveness of the contact-map refinement procedure and other key components, and
(4) robustness under cross-dataset object distribution shifts.

\subsection{Experimental Settings}
\cutsubsectiondown

\paragraph{Train \& Evaluation Settings} 



We train and evaluate all simulation experiments in Isaac Gym~\cite{makoviychuk2021isaac}. For the main benchmark, we use DexGraspNet~\cite{wang2022dexgraspnet} object assets, training on 3,200 objects and evaluating on 141 unseen objects from seen categories and 100 unseen objects from unseen categories, following prior protocols~\cite{xu2023unidexgrasp,wan2023unidexgrasp++,resdex}. Importantly, we use only the object assets and do not rely on ground-truth hand poses. Unless otherwise stated, we use the Shadow Hand in a tabletop lifting task, where success is defined as lifting the object off the table and maintaining a stable hold.

To obtain a vision-based policy under partial observability, we adopt a teacher--student framework following~\cite{wan2023unidexgrasp++}. A privileged state-based policy serves as the teacher, and we distill it into a vision-based student policy using DAgger~\cite{ross2011reduction}. Methods marked with $^\dagger$ denote results reproduced in our work. Additional simulation and training details are provided in Appendix~\ref{app:simulation_training}.

\paragraph{Evaluation Metrics}
We evaluate grasping performance using several complementary metrics.  
Grasp Success Rate (GSR, $\uparrow$) measures success rate of grasp-and-lift trials under a predefined stability criterion.  
Mean Success Affordance Distance (mSAD, $\downarrow$) evaluates how closely the realized contact aligns with the intended region. 
Style Diversity (SD, $\uparrow$) measures the diversity of successful grasp configurations, where higher values indicate more diverse strategies, following prior work~\cite{mao2025universal}.  
We further introduce Intent Success Rate (ISR, $\uparrow$), which counts successful grasps with fingertip contacts within a 4 cm threshold of the intended contact region, providing a stricter measure of intent-consistent lifting.  
Additional implementation details are provided in the Appendix~\ref{sup:evaluation_metrics}.


\vspace{-0.3em}
\begin{table*}[h]
    \centering
    \caption{Grasp Success Rates (GSR, $\uparrow$) on DexGraspNet~\cite{wang2022dexgraspnet} with the Shadow Hand. Our method achieves the highest performance across all settings.}
    \label{tab:main_exp}
    \resizebox{0.98\columnwidth}{!}{
        \begin{tabular}{l|*{3}{c}|*{3}{c}}
            \toprule
            \multicolumn{1}{c|}{\multirow{3}{*}{\textbf{Method}}} &
            \multicolumn{3}{c|}{\textbf{State-Based Setting}} &
            \multicolumn{3}{c}{\textbf{Vision-Based Setting}} \\
            \cmidrule(lr){2-4} \cmidrule(lr){5-7}
            & \multirow{2}{*}{\textbf{Train}} & \multicolumn{2}{c|}{\textbf{Test}}
            & \multirow{2}{*}{\textbf{Train}} & \multicolumn{2}{c}{\textbf{Test}} \\
            \cmidrule(lr){3-4} \cmidrule(lr){6-7}
            & & \textbf{Seen Cat.} & \textbf{Unseen Cat.}
            & & \textbf{Seen Cat.} & \textbf{Unseen Cat.} \\
            \midrule
            UniDexGrasp~\cite{xu2023unidexgrasp} & 79.4 & 74.3 & 70.8 & 73.7 & 68.6 & 65.1 \\
            UniDexGrasp++~\cite{wan2023unidexgrasp++} & 87.9 & 84.3 & 83.1 & 85.4 & 79.6 & 76.7 \\
            UniGraspTransformer~\cite{wang2025unigrasptransformer} & 91.2 & 89.2 & 88.3 & 88.9 & 87.3 & 86.8 \\
            DemoGrasp~\cite{yuan2025demograsp} & 95.2 & 95.5 & 94.4 & 92.2 & 92.3 & 90.1 \\
            $\text{DemoFunGrasp}^\dagger$~\cite{mao2025universal} & 90.0 & 81.7 & 78.7 & 91.4 & 85.8 & 84.5 \\
            \midrule
            \rowcolor{violet!8}
            Ours & \textbf{97.9} & \textbf{98.2} & \textbf{97.7} & \textbf{95.7} & \textbf{95.9} & \textbf{94.9} \\
            \bottomrule
        \end{tabular}
    }
\vspace{-0.2em}
\end{table*}

\subsection{Results on Grasping Performance}
\cutsubsectiondown

Table~\ref{tab:main_exp} compares GSR against existing  dexterous grasping methods on DexGraspNet~\cite{wang2022dexgraspnet}. 
Our method consistently achieves the highest success rates across all splits and settings, reaching 98.2\% on seen categories and 97.7\% on unseen categories under the state-based setting, 
and 95.9\% on seen categories and 94.9\% on unseen categories under the more challenging vision-based setting.
Compared to the strongest prior method, DemoGrasp, our approach improves by 3.5\% and 4.8\% on unseen categories for state-based and vision-based settings, respectively.

DemoGrasp~\cite{yuan2025demograsp} and DemoFunGrasp~\cite{mao2025universal} adapt a fixed demonstration trajectory via demonstration editing to handle diverse objects, while ResDex~\cite{resdex} employs geometry-unaware base policies for generalization. 
While effective, these approaches do not explicitly reason about the local surface geometry of the target contact region when determining hand configurations.
In contrast, our method derives a pseudo target pose from the geometrically refined contact map via inverse kinematics, yielding a hand configuration that is physically grounded in the local structure of each object.
This geometry-aware supervision signal allows the policy to associate similar contact geometries with consistent hand configurations, which we argue underlies the strong generalization to unseen objects.


\subsection{Results on Intent-Aware Grasping}
\label{sec:intent}
\cutsubsectiondown

\paragraph{Intent-Aware Grasp Success Rate}

\begin{wraptable}{r}{0.6\textwidth}
    \vspace{-1.1\baselineskip}
    \centering
    \caption{Comparison of GSR, mSAD, and ISR on~\cite{mao2025universal}. Our method achieves higher success rates while better aligning with intended contact regions.}
    \label{tab:intent_aware}
    \resizebox{0.6\textwidth}{!}{
        \begin{tabular}{l|ccc|ccc}
            \toprule
            \multirow{2}{*}{\textbf{Method}} &
            \multicolumn{3}{c|}{\textbf{Seen Cat.}} &
            \multicolumn{3}{c}{\textbf{Unseen Cat.}} \\
            \cmidrule(lr){2-4} \cmidrule(lr){5-7}
            & \textbf{GSR$\uparrow$} & \textbf{mSAD$\downarrow$} & \textbf{ISR$\uparrow$} 
            & \textbf{GSR$\uparrow$} & \textbf{mSAD$\downarrow$} & \textbf{ISR$\uparrow$} \\
            \midrule
            $\text{DemoGrasp}^\dagger$~\cite{yuan2025demograsp} 
            & 95.50 & 4.10 & 52.57 & 94.40 & 4.27 & 33.71 \\
            $\text{DemoFunGrasp}^\dagger$~\cite{mao2025universal} 
            & 91.92 & 2.92 & 73.74 & 89.90 & 3.23 & 62.63 \\
            \rowcolor{violet!8}
            Ours 
            & \textbf{98.24} & \textbf{2.42} & \textbf{86.52} 
            & \textbf{97.65} & \textbf{2.29} & \textbf{88.87} \\
            \bottomrule
        \end{tabular}
    }
    \vspace{-1.0em}
\end{wraptable}

To evaluate whether successful grasps align with the intended contact regions, we report GSR, mSAD, and ISR. 
GSR measures overall grasp success, mSAD captures the average deviation from the target region, and ISR measures the fraction of successful grasps with fingertip contacts within 4 cm of the intended region.
We reproduce DemoGrasp~\cite{yuan2025demograsp} and DemoFunGrasp~\cite{mao2025universal} under the same evaluation setting using 100 environments.
As shown in Table~\ref{tab:intent_aware}, our method consistently outperforms prior work across both seen and unseen categories. 
In particular, our method achieves higher GSR and ISR, while also obtaining the lowest mSAD of 2.42 cm and 2.29 cm on seen and unseen categories, respectively, improving over DemoFunGrasp by 0.6 cm and 0.92 cm.
This improved contact accuracy leads to higher ISR, demonstrating more accurate grasping of the intended contact regions.
We also observe higher GSR, indicating improved overall grasp success.
Unlike prior methods that rely on a single contact point, our semantically and geometrically grounded contact map provides spatial guidance, enabling more precise localization of the intended contact region during execution.
Additional qualitative visualizations are provided in Appendix~\ref{fig:qual_vis_2}.

\vspace{-0.5mm}
\paragraph{Grasp Style Diversity}
\begin{wraptable}{r}{0.5\textwidth}
    \vspace{-1.0\baselineskip}
    \centering
    \caption{Grasp Style Diversity (SD, $\uparrow$) on ~\cite{wang2022dexgraspnet}. Our method produces more diverse successful grasp strategies.}
    \label{tab:style_diversity}
    \resizebox{0.48\textwidth}{!}{
        \begin{tabular}{lcc}
        \toprule
        \textbf{Method}  & \textbf{Seen Cat.} & \textbf{Unseen Cat.} \\
        \midrule
        $\text{DemoGrasp}^\dagger$\cite{yuan2025demograsp}   & 1.66 & 1.65 \\
        DemoFunGrasp\cite{mao2025universal}   & 1.48 & 1.44 \\
        \rowcolor{violet!8}
        Ours  & \textbf{1.71} & \textbf{1.80} \\
    \bottomrule
    \end{tabular}
    }
\end{wraptable}
Since each object has multiple semantic intentions, the policy should generate distinct grasp strategies rather than collapsing to a single strategy.
A policy that fails to account for such variation would produce similar grasp configurations across different situations.
To evaluate whether the policy avoids such collapse, we measure grasp style diversity (SD) on the DexGraspNet~\cite{wang2022dexgraspnet}, which quantifies the diversity of successful grasping strategies by computing the average Euclidean distance between all pairs of successful hand joint configurations.
As shown in Table~\ref{tab:style_diversity}, our method achieves higher SD than prior works~\cite{xu2023unidexgrasp, yuan2025demograsp, mao2025universal}.
Importantly, this improvement comes alongside higher grasp success rates, indicating that the gains are not due to trivial diversification but reflect effective grasp strategies.

\subsection{Qualitative Results}
\cutsubsectiondown

\begin{figure}[t]
  \centering
    \vspace{-1mm}
    \includegraphics[width=0.95\textwidth]{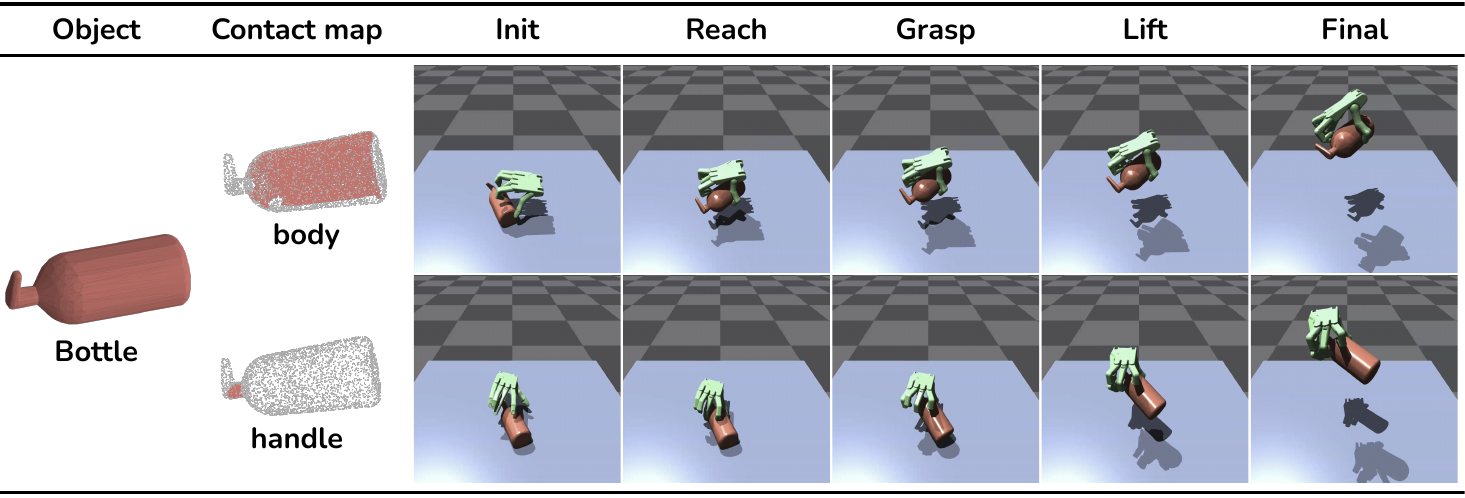}
      \caption{\small Qualitative results of intent-aware grasping. Given different intents on the same object, SECOND-Grasp adapts its grasp behavior to the specified contact region.}
  \label{fig:intent_vis}
\vspace{-1.0mm}
\end{figure}

\begin{figure}[t]
  \centering
\vspace{-1mm}
  \includegraphics[width=0.98\textwidth]{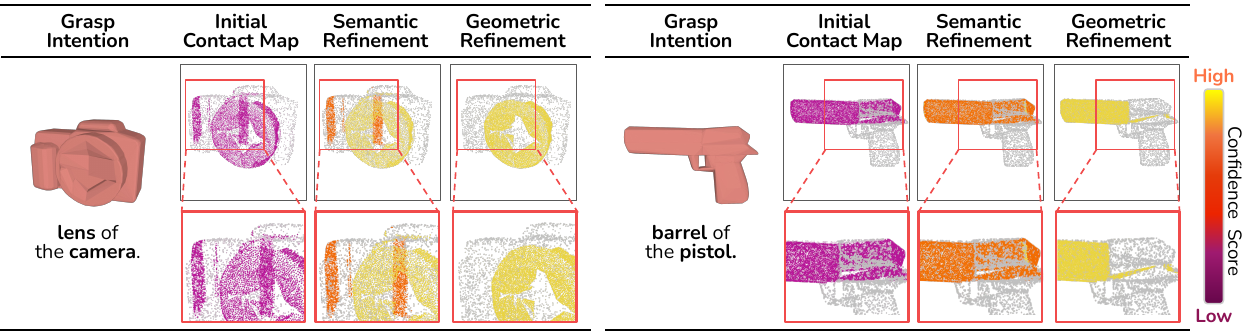}
  \caption{\small Qualitative visualization of SGCR. 
  Given a grasp intention, SGCR refines coarse contact proposals into localized and geometrically coherent contact maps. Warmer colors indicate higher contact confidence.}
    \vspace{-1mm}
  \label{fig:sgcr_vis}
\end{figure}

We present qualitative results of the proposed method.
Figure~\ref{fig:sgcr_vis} visualizes the refinement process of Semantic-Geometric Consistency Refinement (SGCR), where coarse semantic contact proposals are refined into more localized and geometrically consistent contact regions on the object surface. 
Figure~\ref{fig:intent_vis} shows grasp executions conditioned on different contact maps. 
For the same object, specified regions lead to distinct grasp behaviors, such as grasping the bottle body when highlighted and grasping the handle when specified. 
These results show that our method follows the given contact intent rather than collapsing to a fixed grasp strategy. 
See Appendix~\ref{app:add_result} for more examples.

\subsection{Ablation Study}
\cutsubsectiondown

\begin{wraptable}{r}{0.56\textwidth}
    \vspace{-1.0\baselineskip}
    \centering
    \caption{Ablation study on pseudo pose, contact map, and SGCR, evaluated using GSR and SD. Each component consistently improves grasp success and diversity.}
    \label{tab:ablation}
    \resizebox{0.56\textwidth}{!}{
        \large  
        \begin{tabular}{ccc|cc|cc}
            \toprule
            \multirow{2}{*}{\textbf{Pseudo Pose}} &
            \multirow{2}{*}{\textbf{Contact Map}} &
            \multirow{2}{*}{\textbf{SGCR}} &
            \multicolumn{2}{c|}{\textbf{Seen Cat.}} &
            \multicolumn{2}{c}{\textbf{Unseen Cat.}} \\
            \cmidrule(lr){4-5} \cmidrule(lr){6-7}
            & & &
            \textbf{GSR$\uparrow$} & \textbf{SD$\uparrow$} &
            \textbf{GSR$\uparrow$} & \textbf{SD$\uparrow$} \\
            \midrule
            \xmark & \xmark & \xmark & 90.82 & 0.90 & 88.96 & 0.89 \\
            \cmark & \xmark & \xmark & 93.77 & 1.27 & 92.77 & 1.21 \\
            \cmark & \cmark & \xmark & 97.85 & 1.54 & 96.88 & 1.68 \\
            \midrule
            \rowcolor{violet!8}
            \cmark & \cmark & \cmark &
            \textbf{98.24} & \textbf{1.71} &
            \textbf{97.65} & \textbf{1.80} \\
            \bottomrule
        \end{tabular}
    }
\end{wraptable}

We conduct an ablation study to analyze each component, including the pseudo pose prior, contact map, and Semantic-Geometric Consistency Refinement(SGCR).
The pseudo pose prior serves as an alternative to dataset-provided joint configurations, maintaining performance while alleviating limited diversity of predefined grasp poses.
However, without the contact map, the pseudo pose prior alone cannot generate grasps that reflect object-specific semantics or geometry.
The contact map provides spatial cues that guide where to make contact, enabling semantically diverse grasp generation.
Finally, SGCR refines the contact map through semantically grounded geometric consistency, improving grasp accuracy while preserving diversity.
As SGCR refines the contact map, it is evaluated with the contact map.


\subsection{Cross-Dataset Generalization}
\cutsubsectiondown

Table~\ref{tab:main_exp} shows that our method achieves strong performance on DexGraspNet~\cite{wang2022dexgraspnet}. 
To assess robustness under distribution shift, we further evaluate cross-dataset performance on several object datasets, including DexGraspAnything(DGA)~\cite{zhong2025dexgrasp}, EGAD~\cite{morrison2020egad}, Omni6DPose~\cite{zhang2024omni6dpose}, ModelNet40~\cite{wu20153d}, and VisualDexterity~\cite{chen2023visual}(Table~\ref{tab:x_data}).
Our method consistently achieves the highest success rates across datasets. 
Notably, even compared to DemoGrasp~\cite{yuan2025demograsp}, which additionally trains on YCB~\cite{calli2015ycb}, our method  substantial improvements on DGA (+18.81\%) and ModelNet40 (+16.01\%), indicating stronger robustness under distribution shift.
We attribute this to our geometrically grounded contact maps, derived directly from object geometry and provide stable guidance across different objects.
For a fair comparison, we follow prior work~\cite{yuan2025demograsp} and randomly rescale object sizes to 6–15 cm.

\begin{table}[h]
    \vspace{-1.0mm}
    \caption{Cross-dataset generalization (GSR$, \uparrow$) on five object datasets. Our method demonstrates improved robustness to distribution shifts.}
    \label{tab:x_data}
    \centering
    \resizebox{\textwidth}{!}{
    \begin{tabular}{l|c|ccccc}
    \toprule
    \textbf{Method} & \diagbox{\textbf{Train}}{\textbf{Test}} 
    & \textbf{DGA} & \textbf{EGAD} & \textbf{Omni6DPose} & \textbf{ModelNet40} & \textbf{VisualDexterity} \\
    \midrule
    $\text{UniGraspTransformer}^\dagger$~\cite{wang2025unigrasptransformer} & DexGraspNet & 44.53 & 69.80 & 47.95 & 40.45 & 68.63 \\
    DemoGrasp\cite{yuan2025demograsp} & DexGraspNet + YCB & 60.99 & 94.58 & 74.84 & 71.88 & 93.67 \\
    $\text{DemoFunGrasp}^\dagger$\cite{mao2025universal} & DexGraspNet + YCB & 50.00 & 83.84 & 57.58 & 50.50 & 81.82 \\
    \rowcolor{violet!8}
    Ours & DexGraspNet 
    & \textbf{79.80} & \textbf{96.68} & \textbf{77.84} & \textbf{87.89} & \textbf{94.14} \\
    \bottomrule
    \end{tabular}}
    \vspace{-0.7mm}
\end{table}
\section{Conclusion}
\cutsectiondown


We presented SECOND-Grasp, a framework for dexterous grasping that bridges semantic intent and physical feasibility through contact-level reasoning.
By combining vision-language reasoning, multi-view segmentation, and Semantic-Geometric Consistency Refinement, our method produces reliable 3D contact maps.
From these contact maps, we derive pseudo target poses via inverse kinematics, providing pose priors for object-specific and physically feasible policy learning.
SECOND-Grasp further derives contact maps and pose priors without training, enabling generalization to unseen objects.
Extensive experiments demonstrate that our method improves grasping performance while capturing semantic intent and ensuring stable lifting.
Results across multiple cross-dataset benchmarks confirm generalization, and comprehensive ablations validate the effectiveness of our approach.

\textbf{Limitations and Future work.}
Our experiments focus on dexterous hand manipulation without explicitly evaluating the proposed method on an arm-hand robotic platform. 
This choice was motivated by the fact that dexterous hands are high-DoF and inherently complex, making them challenging. 
We believe that demonstrating reliable performance at the hand level is a necessary first step toward full-body or arm-hand manipulation. 
Our results show that the proposed method can effectively handle this challenging setting, providing evidence that it can serve as a foundation for more complete robotic systems. 
In future work, we plan to extend our framework to arm-hand platforms and evaluate coordinated control.

\clearpage
\bibliographystyle{abbrv}
\bibliography{ref}

\newpage
\appendix

\section{Additional Implementation Details}
\cutsectiondown

\subsection{Simulation and Training Setup}
\label{app:simulation_training}

We conduct our experiments in Isaac Gym~\cite{makoviychuk2021isaac} using a dexterous hand in a tabletop manipulation environment.
The table size is set to $1.0\times1.0\times0.6$ m, and objects are initialized at a height of $0.7$ m, i.e., $0.1$ m above the tabletop.
For the experiments, we use 
DexGraspNet~\cite{wang2022dexgraspnet} objects with scale factors of \(\{0.06,0.08,0.10,0.12,0.15\}\).
The task goal is to lift an object placed on the table to $0.3$ m above the tabletop.
Each training run takes approximately 16 hours on a single NVIDIA GeForce RTX 4090 GPU.
The training hyperparameters used in simulation are reported in Table~\ref{tab:policy_hyperparams}.

\begin{table}[h]
\vspace{-1.0mm}
\centering
\caption{Hyperparameters of our experimental setup.}
\label{tab:policy_hyperparams}
\footnotesize
\setlength{\tabcolsep}{6pt}
\renewcommand{\arraystretch}{1.15}
\begin{tabularx}{0.58\linewidth}{@{}X r@{}}
\toprule[1.2pt]
\textbf{Hyperparameter} & \textbf{Value} \\
\midrule[0.7pt]
\rowcolor{gray!10}[0pt][0pt]
\multicolumn{2}{@{}l@{}}{\textbf{Common}} \\
Episode length & 200 \\
Training epochs per iteration & 5 \\
Action moving average coefficient & 1.0 \\
Training iterations & 20{,}000 \\
Random seed & 8 \\
\addlinespace[2pt]
\rowcolor{gray!10}[0pt][0pt]
\multicolumn{2}{@{}l@{}}{\textbf{State-based policy}} \\
Number of environments & 11{,}000 \\
Parallel rollout steps per iteration & 8 \\
Number of minibatches per epoch & 8 \\
\addlinespace[2pt]
\rowcolor{gray!10}[0pt][0pt]
\multicolumn{2}{@{}l@{}}{\textbf{Vision-based policy}} \\
Number of environments & 4{,}096 \\
Parallel rollout steps per iteration & 1 \\
Number of minibatches per epoch & 4 \\
\bottomrule[1.2pt]
\end{tabularx}
\vspace{-0.3mm}
\end{table}


\subsection{Reward Terms}

\label{app:reward_details}
Here, we provide additional implementation details for the reward terms in Eq.~\ref{eq:policy_reward}.
In implementation, each reward term is computed at every simulation timestep $t$.
The policy reward consists of three components: a pose-guidance reward, a contact-conditioned pose reward, and the task reward. 
The first two terms encourage the hand to follow the reference hand state derived from our pseudo pose prior $\theta^*$, and contact map $\mathcal{Q}_{\mathrm{final}}^{(k)}$, while the task reward drives the actual grasp-and-lift behavior.
We first define a pose-tracking score that measures how closely the current hand state matches the reference hand state:
\begin{equation}
    r_t^{\mathrm{track}}
    =
    \exp\!\left(
    -\lambda_w \|\Delta w_t\|_2
    -\lambda_\phi \|\Delta \phi_t\|_2
    -\lambda_\theta \|\Delta \theta_t\|_1
    \right),
\label{eq:track}
\end{equation}
where \(\Delta w_t\), \(\Delta \phi_t\), and \(\Delta \theta_t\) denote the wrist-position, wrist-orientation, and joint-configuration errors with respect to the reference hand state, respectively. 
We set \(\lambda_w=7.0\), \(\lambda_\phi=2.0\), and \(\lambda_\theta=0.12\).
The pose-guidance reward is then defined as
\begin{equation}
r_t^{\mathrm{pose}}
=
\beta\,\kappa_t\,r_t^{\mathrm{track}},
\label{eq:rpose}
\end{equation}
where \(\beta=0.55\). 
The weight \(\kappa_t\) decays over the first 80 time steps with a minimum value of \(0.15\).
This emphasizes reference-pose guidance early in the episode and gradually reduces its effect over time.
To encourage the fingertips to reach the intended contact region \(\mathcal{Q}_{\mathrm{final}}^{(k)}\), we add a contact-conditioned pose reward:
\begin{equation}
r_t^{\mathrm{contact}}
=
\beta_c\,b_t\,r_t^{\mathrm{track}},
\label{eq:rcontact}
\end{equation}
Here, \(b_t\in\{0,1\}\) denotes a binary indicator of fingertip contact with \(\mathcal{Q}_{\mathrm{final}}^{(k)}\), and we set \(\beta_c=0.25\).
In addition to the pose-guidance terms, we use a task reward \(r_t^{\mathrm{task}}\) based on the grasp-and-lift reward from the baseline~\cite{resdex}. 
It consists of approach, lift, goal-reaching, and bonus terms:
\begin{equation}
r_t^{\mathrm{task}}
=
r_t^{\mathrm{approach}}
+
r_t^{\mathrm{lift}}
+
r_t^{\mathrm{goal}}
+
r_t^{\mathrm{bonus}}.
\end{equation}

The approach term guides the hand toward the object.
The lift term encourages the object to be raised from the table, and the goal-reaching term encourages the lifted object to move toward the target position.
The bonus term provides an additional reward when the object is sufficiently close to the target.
The total reward at timestep \(t\) is given by
\begin{equation}
r_t = r_t^{\mathrm{pose}} + r_t^{\mathrm{contact}} + r_t^{\mathrm{task}}.
\end{equation}

\subsection{Evaluation Metrics}
\label{sup:evaluation_metrics}
We evaluate each method using the following four metrics.
\begin{itemize}
    \item \textbf{Grasp Success Rate (GSR)}  
    GSR measures the fraction of episodes in which the object is lifted and held near the target position.
    An episode is considered successful if the object remains within \(5\) cm of the lifting goal for \(20\) consecutive time steps.
    \item \textbf{Mean Success Affordance Distance (mSAD)}  
    mSAD measures intent alignment using the Euclidean distance between fingertip contacts and the intended contact region.
    It is averaged over successful episodes only, with lower values indicating better alignment.
    \item \textbf{Style Diversity (SD)}  
    SD measures the diversity of hand postures among successful episodes.
    We collect the final hand joint configuration from each successful episode, compute pairwise Euclidean distances among them, and report the average distance.
    \item \textbf{Intent Success Rate (ISR)}  
    ISR measures the fraction of episodes that satisfy both grasp success and intent-consistent contact.
    For episode \(n\), intent success is defined as \(\mathrm{GSR}_n=1\) and \(\mathrm{SAD}_n < 0.04\,\mathrm{m}\).
    The reported ISR is averaged over all evaluation episodes.
\end{itemize}


\section{Contact Map Generation Details}
\cutsectiondown

\subsection{VLM-based Grasp Intent Proposal}
\label{app:vlm_intent_proposal}

For each object, we render four side-view RGB images (e.g., front, left, back, and right) and provide them jointly to Qwen3-VL-8B-Instruct~\cite{bai2025qwen3}. 
The VLM predicts between two and four grasp intents, where each intent corresponds to a distinct object-centric graspable part.

\begin{figure}[t]
  \centering
  \includegraphics[width=\textwidth]{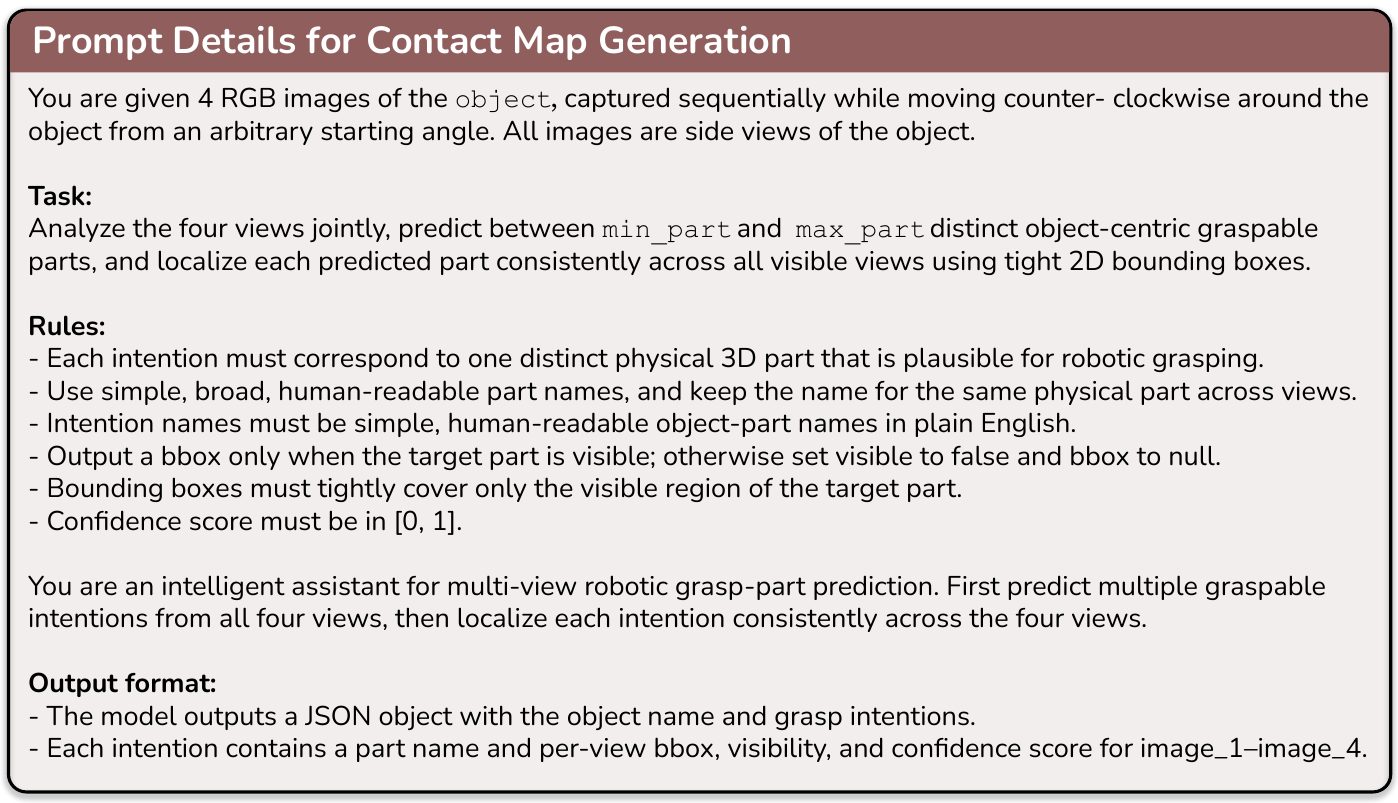}
    \caption{Prompt used for VLM-based grasp intent proposal, including the JSON output format.}
  \label{fig:prompt_vis_2}
\vspace{-1.0mm}
\end{figure}

For each intent, the VLM outputs a part name, a short description, per-view visibility, a 2D bounding box for each visible view, and a confidence score. 
The prompt discourages overly complex object-part names, enforces consistent naming across views, and suppresses bounding-box predictions for views where the part is not visible. 
The VLM output is required to be valid JSON, which is parsed into a structured set of per-view part proposals.
The full prompt, including a compact description of the required JSON output format, is shown in Fig.~\ref{fig:prompt_vis_2}.

We decode with temperature \(0.2\), repetition penalty \(1.1\), and a maximum of \(3000\) new tokens. 
During post-processing, invalid bounding boxes are removed and confidence scores are clipped to \([0,1]\). 
If the visibility flag is missing or unreliable, we determine visibility based on whether a valid bounding box exists.
These predictions are used as coarse semantic proposals; the final 3D contact maps are obtained through mask extraction, depth projection, and Semantic-Geometric Consistency Refinement.

\subsection{Semantic-Geometric Contact Map Generation}
\label{app:contact_map_generation}

This section provides additional implementation details for generating the semantic-geometric contact maps used in our experiments. 
Starting from the VLM-predicted~\cite{bai2025qwen3} per-view bounding boxes, we first apply SAM2~\cite{ravi2024sam} within each box to obtain a binary mask for the corresponding graspable part. 
The extracted mask is intersected with the valid object silhouette obtained from the rendered depth map, which removes mask leakage outside the object region.

For each grasp intention, we then construct a per-view confidence map by assigning the VLM confidence score to pixels inside the filtered mask and zero elsewhere. 
These confidence maps are refined using neighboring-view support: mask pixels are lifted to 3D using the rendered depth and camera parameters, reprojected to adjacent views, and considered supported only when the reprojected location is depth-consistent and lies inside the corresponding part Eq.~\ref{eq:depth_consistency}. 
Supported pixels receive additional confidence from the neighboring view, and the resulting maps are globally normalized across the four views.

The normalized 2D confidence maps are finally lifted to the object surface to obtain a scored 3D point cloud for each grasp intention. 
We aggregate points from all visible views and associate each point with its refined confidence score. 
To remove fragmented or geometrically inconsistent regions, we select a high-confidence seed component and expand it only to nearby candidate points that satisfy the local convexity criterion in Eq.~\ref{eq:local_convex_pair}. 
The remaining points form the final semantic-geometric contact map used for inverse-kinematics-based pose initialization and policy learning. 
Figure~\ref{fig:sgcr_vis} visually illustrates the overall semantic-geometric contact map generation process.

\begin{figure}[t]
\vspace{-1.0mm}
  \centering
  \includegraphics[width=\textwidth]{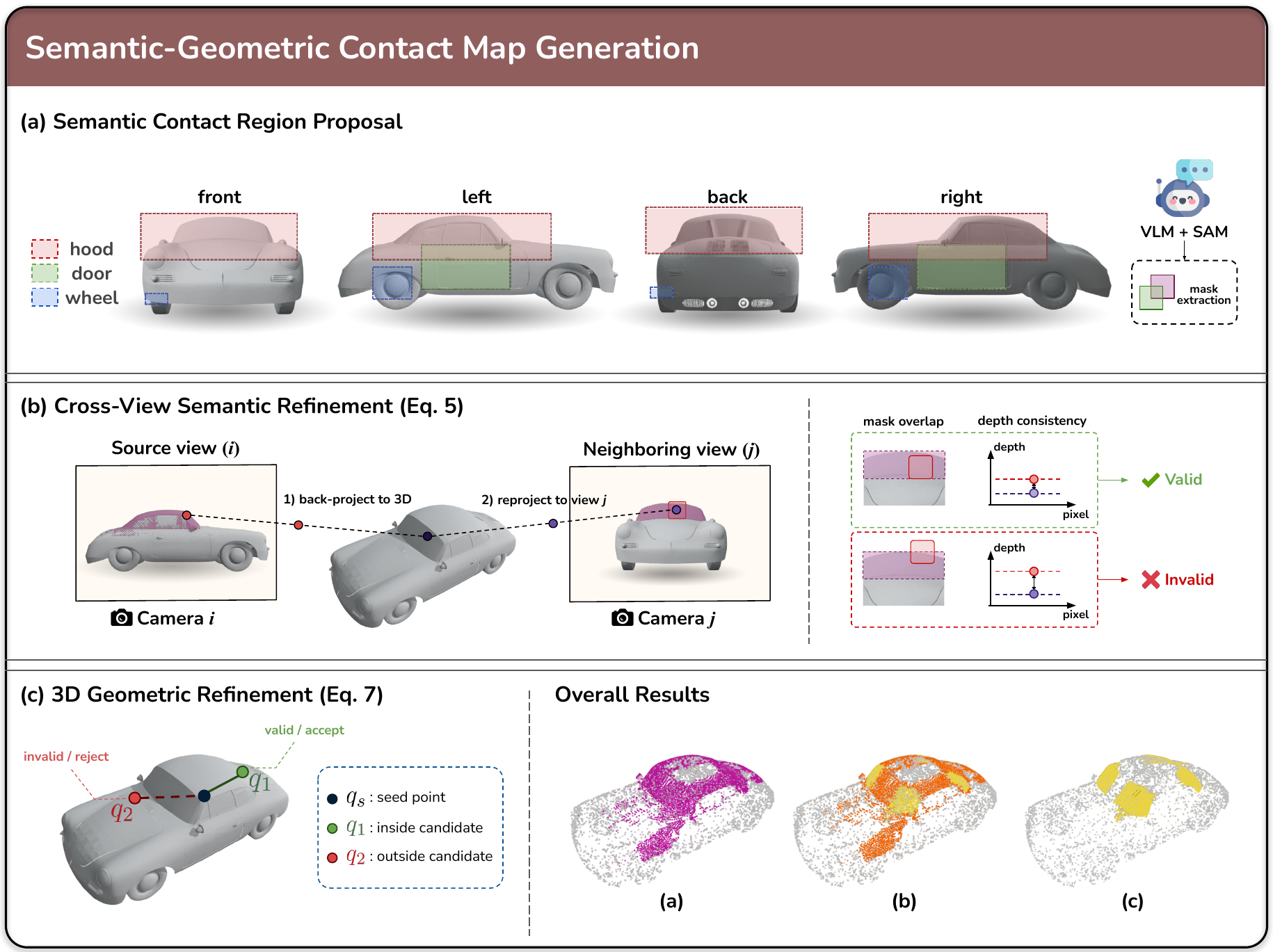}
  \caption{Overview of Semantic Contact Map Generation. Visual illustration of the geometric refinement steps in Semantic-Geometric Consistency Refinement.}
  \label{fig:sgcr_vis}
\vspace{-5.0mm}
\end{figure}


\section{Inverse Kinematics}
\cutsectiondown

\subsection{Jacobian-Based Inverse Kinematics}
\label{app:ik_solver}

This appendix details the local damped least-squares solver~\cite{buss2004introduction} used to optimize Eq.~\ref{eq:ik_objective}. 
Starting from an initial hand configuration $h_0=(w_0,\phi_0,\theta_0)$, we perform 12 iterative updates.

At each update, given the current configuration \(h\), we first instantiate the closest target point for each fingertip within its assigned contact region:
\begin{equation}
    y_f^{\star}
    =
    \arg\min_{y\in \mathcal{R}_f^{(k)}}
    \left\|
    \xi_f(h)-y
    \right\|_2 .
\end{equation}
The fingertip-position residuals are then concatenated as
\begin{equation}
    e =
    \left[
    \xi_f(h) - y_f^{\star}
    \right]_{f\in F}.
\end{equation}
Let \(J=\partial e / \partial h\) denote the stacked fingertip Jacobian evaluated at the current configuration. 
We compute the damped least-squares update as
\begin{equation}
    \Delta h
    =
    -J^\top
    \left(
    JJ^\top+\lambda_{\mathrm{dls}}^2 I
    \right)^{-1}
    e,
    \qquad
    h
    \leftarrow
    h+\eta\Delta h,
\end{equation}
where \(\lambda_{\mathrm{dls}}\) is the damping coefficient and \(\eta\) is the step size. 
After each update, the nearest target points are recomputed from the assigned contact regions, yielding a local alternating procedure between target-point selection and IK update.

Let the final IK state be \(h^\star=(w^\star,\phi^\star,\theta^\star)\). 
During training, we use only the joint component \(\theta^\star\) as pseudo-pose supervision. 
The IK result is used only as a pseudo-pose label for training, not as a robot trajectory, and is not used during evaluation.


\subsection{Analysis of IK-based Pseudo Pose Guidance}

The role of the pseudo pose in SECOND-Grasp is not to produce a final executable grasp, but to provide an intermediate hand-level prior for policy learning. 
Given a refined semantic-geometric contact map, we only need a hand configuration that roughly aligns the fingertips with the intended contact region while respecting the kinematic structure of the target hand. 
Thus, we prioritize efficient contact-aligned guidance over exhaustive pose optimization.

To determine the number of IK iterations used in our pseudo-pose generation pipeline, we evaluate the efficiency--alignment trade-off by running IK with 1, 3, 12, and 50 iterations and comparing it with baseline pose-generation methods on the same refined contact maps \(\mathcal{Q}_{\mathrm{final}}^{(k)}\). 
We sample 100 objects from DexGraspNet~\cite{wang2022dexgraspnet} and report inference time and Contact Coverage (Cov.). 
Inference time is measured in seconds as the average wall-clock time required to generate one pseudo pose from the same object and refined contact map input.
Following prior works~\cite{wei2024grasp, ma2025contact}, Cov.@\(\tau\) measures the percentage of hand surface points whose signed distance to the object surface lies within \([-\tau,\tau]\), where \(\tau \in \{2\text{ mm},5\text{ mm}\}\).

As shown in Table~\ref{tab:app_pseudo_pose_comparison}, increasing the number of IK iterations improves contact alignment.
The 12-iteration setting achieves the best Cov.@2mm and matches the best Cov.@5mm among IK settings, while increasing to 50 iterations does not improve coverage and increases inference time by more than \(4\times\). 
Based on this trade-off, we adopt 12 IK iterations throughout our method. 
Compared with optimization- and generative-based pose-generation methods, this setting provides competitive contact alignment with substantially lower computational cost. 
These results support our choice of IK-based pseudo poses as efficient intermediate guidance for policy learning rather than final grasp synthesis.

\begin{table}[h]
\vspace{-5mm}
\centering
\renewcommand{\arraystretch}{1.05}
\caption{Efficiency and contact-alignment comparison for pseudo-pose generation. IK with 12 iterations provides a practical trade-off between contact coverage and inference time, making it suitable as intermediate guidance for policy learning.}
\label{tab:app_pseudo_pose_comparison}
\resizebox{\linewidth}{!}{%
\begin{tabular}{lccccccc}
\toprule
\multirow{2}{*}{Metrics}
& \multicolumn{4}{c}{Inverse Kinematics}
& \multicolumn{1}{c}{Optimization-based Method}
& \multicolumn{2}{c}{Generative-based Method} \\
\cmidrule(lr){2-5} \cmidrule(lr){6-6} \cmidrule(lr){7-8}
& Iter. $=1$
& Iter. $=3$
& Iter. $=12$
& Iter. $=50$
& GenDexGrasp~\cite{li2023gendexgrasp}
& UGG~\cite{lu2024ugg}
& DGA~\cite{zhong2025dexgrasp} \\
\midrule
Inf. Time (s) $\downarrow$
& 0.004 & 0.037 & 0.187 & 0.836 & 11.329 & 9.222 & 0.622  \\
Cov.@2mm (\%) $\uparrow$
& 2.220 & 3.330 & 3.780 & 3.110 & 2.622 & 0.001 & 1.206  \\
Cov.@5mm (\%) $\uparrow$
& 12.000 & 15.110 & 15.110 & 14.440 & 16.585 & 2.547 & 8.839  \\
\bottomrule
\end{tabular}%
}
\vspace{-4mm}
\end{table}

\subsection{Effect of IK-derived Pseudo Pose on Policy Learning}
\label{app:ik_policy_learning}

We further analyze how the IK-derived pseudo pose affects policy learning. 
Learning dexterous grasping from task reward alone often requires extensive exploration, as the policy must discover effective contact locations and finger coordination by itself.
The IK-derived pseudo pose converts the refined contact map into a joint-level target \(\theta^\star\). 
This provides the policy with a feasible hand configuration whose fingertips are already close to the intended contact regions.

\begin{figure}[h]
  \centering
  \includegraphics[width=0.6\textwidth]{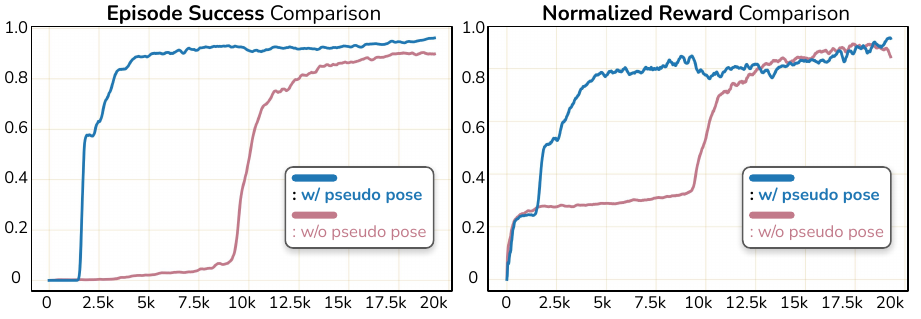}
  \vspace{-2mm}
  \caption{Effect of IK-derived Pseudo-Pose Supervision on Policy Learning.}
  \label{fig:plot_supple_v2}
  \vspace{-2mm}
\end{figure}

To isolate this effect, we train two policy variants under the same task reward, object split, simulation setting, and hyperparameters: one with auxiliary supervision toward the pseudo pose \(\theta^\star\), and one without it. 
For each variant, we plot the training time episode success rate and normalized reward in Figure~\ref{fig:plot_supple_v2}. 
Without pseudo-pose supervision, the policy improves only after a long delay, indicating that contact discovery from task feedback alone is inefficient. 
In contrast, the policy trained with \(\theta^\star\) achieves high reward and success much earlier, showing that the pseudo pose serves as an intermediate hand-pose prior that accelerates early contact formation and grasp stabilization. 
The final performance gap is smaller than the early training gap, suggesting that the main benefit of pseudo-pose supervision is not to prescribe the final grasp, but to guide the policy toward contact-rich states during learning.

\section{Additional Results}
\label{app:add_result}
\cutsectiondown

\subsection{Transfer to Allegro Hand}
\label{app:transfer_allegro_hand}
We further test whether the same contact-map representation can support grasping with a different robot hand, the Allegro Hand.
This transfer is not a direct pose copy because the two hands have different finger layouts and joint spaces.
Shadow Hand commonly has five digits and 20 actuated DoFs, whereas Allegro Hand has four fingers and 16 actuated DoFs.
Therefore, a joint pose defined for Shadow Hand cannot be directly applied to Allegro Hand.
Prior methods~\cite{yuan2024cross} often transfer grasps through hand-specific pose retargeting, which may require additional optimization or training.
Our method instead keeps the contact map on the object and recomputes the joint target using IK with the Allegro Hand.
Thus, the intended object region stays the same, while the hand pose is adapted to the new hand embodiment.

\begin{figure}[h]
  \vspace{-5mm}
  \centering
  \includegraphics[width=\textwidth]{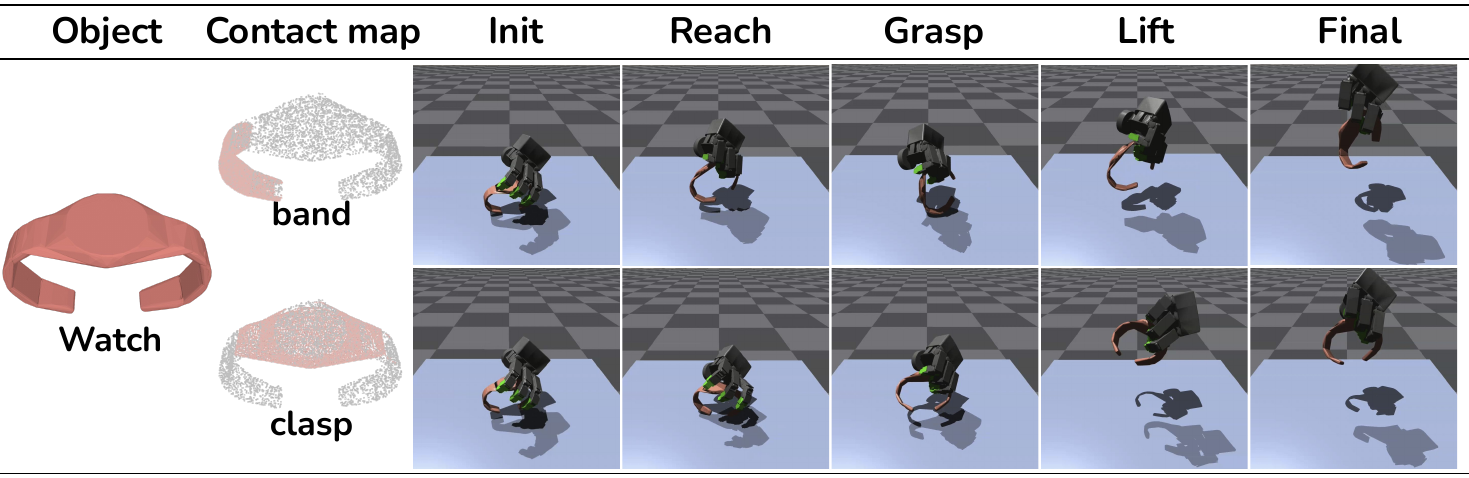}
\caption{Additional qualitative results of intent-aware grasping on the Allegro Hand.}
  \label{fig:allegro_qual_vis}
  \vspace{-2mm}
\end{figure}

Figure~\ref{fig:allegro_qual_vis} shows that the Allegro Hand grasps and lifts the object from the intended region, such as the watch band or clasp.
Table~\ref{tab:allegro_results} shows that 12 iterations of IK provide a good balance between runtime and contact coverage.
Compared with the pose-generation baseline, our result is much faster \((0.302\,\mathrm{s}\) vs. \(5.086\,\mathrm{s})\) and achieves higher Cov.@5mm \((0.158\) vs. \(0.033)\).
These results show that our method provides reliable contact map and effective initial poses even for a different hand embodiment, demonstrating that the intent-aware representation can transfer beyond the original hand embodiment.

\begin{table}[h]
\centering
\renewcommand{\arraystretch}{1.05}
\caption{Comparison of pseudo-pose efficiency across different IK iterations and baseline pose-generation methods. All methods are evaluated on the same sampled DexGraspNet training objects.}
\label{tab:allegro_results}
\resizebox{\linewidth}{!}{%
\begin{tabular}{llccccc}
\toprule
\multirow{2}{*}{Hand / Dataset}
& \multirow{2}{*}{Metrics}
& \multicolumn{4}{c}{Inverse Kinematics}
& \multicolumn{1}{c}{Generative-based Method}\\
\cmidrule(lr){3-6} \cmidrule(lr){7-7} 
&
& Iter. $=1$
& Iter. $=3$
& Iter. $=12$
& Iter. $=50$
& Dexdiffuser~\cite{weng2024dexdiffuser}\\
\midrule
\multirow{3}{*}{\begin{tabular}[c]{@{}l@{}}Allegrohand\\(DexGraspNet)\end{tabular}}
& Inf. Time (s) $\downarrow$
& 0.028 & 0.078 & 0.302 & 0.468 & 5.086  \\
& Cov.@2mm $\uparrow$
& 0.005 & 0.026 & 0.038 & 0.040 & 0.003  \\
& Cov.@5mm $\uparrow$
& 0.056 & 0.087 & 0.158 & 0.173 & 0.033  \\
\bottomrule
\end{tabular}%
}
\end{table}

\subsection{Qualitative Results of Contact-Map Refinement}

\begin{figure}[t]
  \centering
  \includegraphics[width=\textwidth]{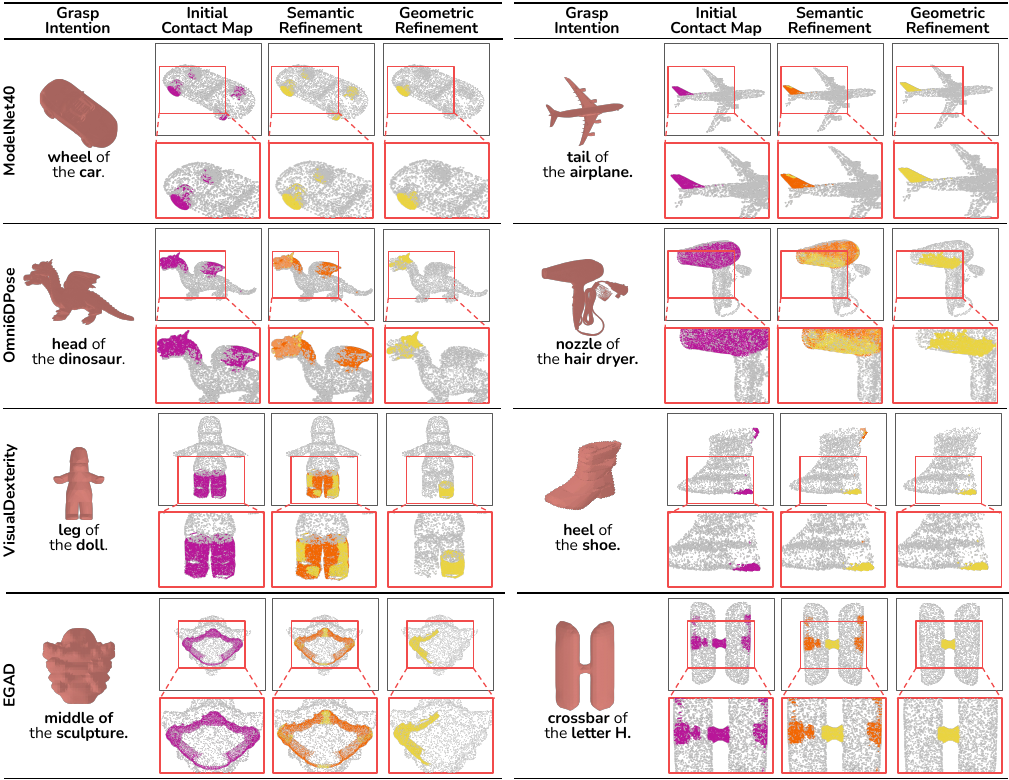}
\caption{Cross-dataset qualitative results of SGCR contact-map refinement. Coarse contact proposals are refined into geometrically coherent contact maps on objects from diverse datasets.}
  \vspace{-5mm}
  \label{fig:sgcr_supple}
\end{figure}

Figure~\ref{fig:sgcr_supple} shows additional qualitative examples of Semantic-Geometric Consistency Refinement (SGCR) on objects from datasets beyond DexGraspNet~\cite{wang2022dexgraspnet}. 
Starting from coarse contact proposals, SGCR removes noisy regions and refines the contact map around the intent-specified object part.
Across diverse datasets, including DexGraspAnything (DGA)~\cite{zhong2025dexgrasp}, EGAD~\cite{morrison2020egad}, Omni6DPose~\cite{zhang2024omni6dpose}, ModelNet40~\cite{wu20153d}, and VisualDexterity~\cite{chen2023visual}, the refined maps become more compact and geometrically consistent while preserving the intended contact semantics. 
These results provide additional evidence that SGCR remains effective under cross-dataset object distribution shifts.

\subsection{Qualitative Results of Intent-aware Grasping}
\label{app:qualitative_results}
\begin{figure}[t]
  \centering
  \includegraphics[width=\textwidth]{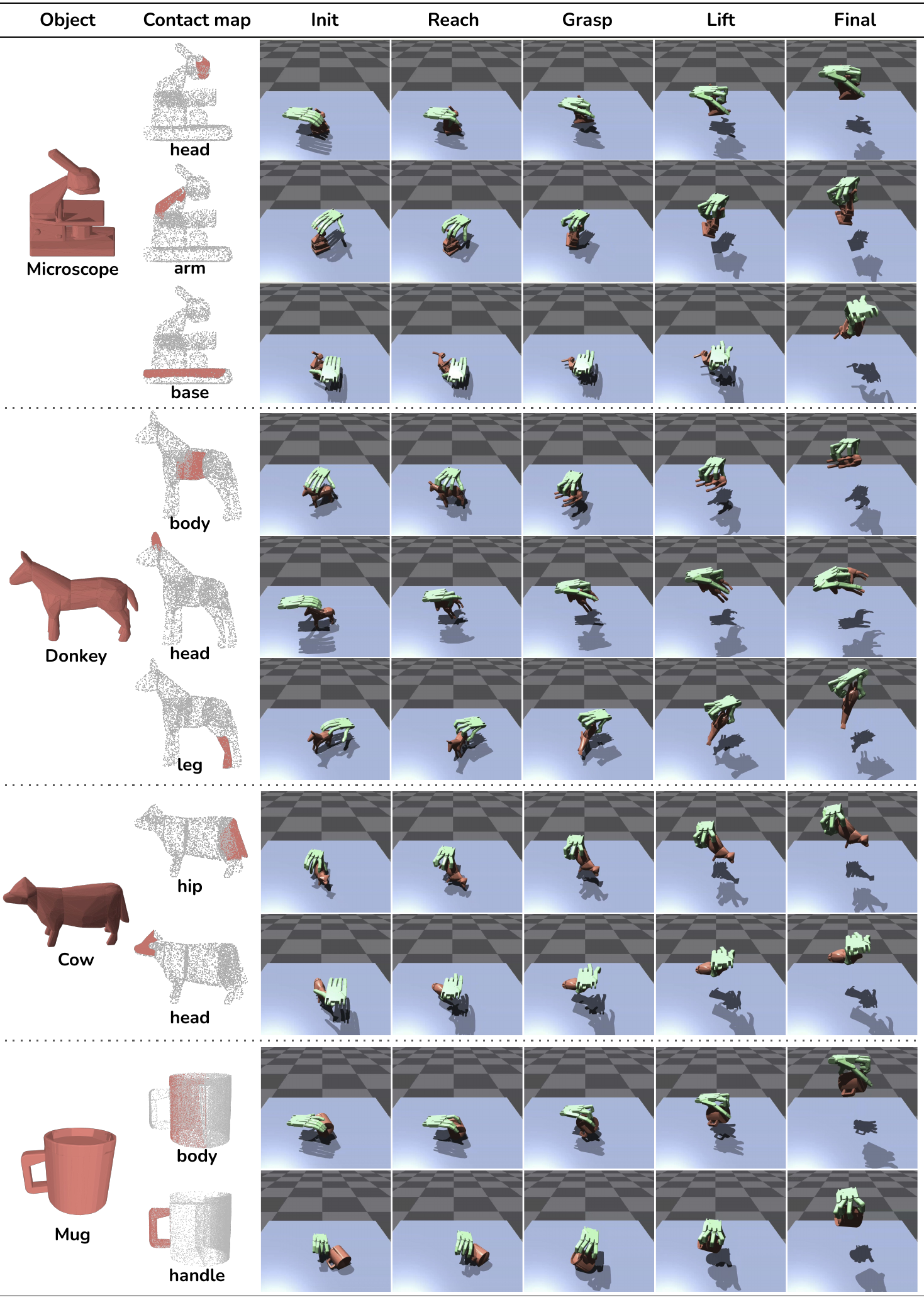}
\caption{Additional qualitative results of intent-aware grasping. SECOND-Grasp adapts the refined contact map and resulting grasp behavior according to the specified semantic part intent.}
  \vspace{-5mm}
  \label{fig:qual_vis_2}
\end{figure}

Figure~\ref{fig:qual_vis_2} presents additional qualitative examples of intent-aware grasping across diverse object categories and semantic part intents.
Rather than producing a single stereotyped grasp, our method adapts the refined contact map and the resulting grasp behavior to the requested intent.
For instance, on the same microscope object, different intents such as head, arm, and base lead to distinct contact maps concentrated on the corresponding semantic regions.
The learned policy then follows these intent-specific contact cues to approach, grasp, and lift the object from the desired part, resulting in clearly different hand placements.
Across objects and semantic parts, these examples highlight that our method performs intent-conditioned grasping instead of relying on a fixed grasping pattern.

\clearpage
\newpage

\end{document}